# Multi-robot Cooperative Box-pushing problem using Multi-objective Particle Swarm Optimization Technique

Arnab Ghosh, Avishek Ghosh, Arkabandhu Chowdhury, Amit Konar

Dept. of Electronics & Telecommunication Engineering
Jadavpur University
Kolkata 700032, India
arnabju90@gmail.com, avishek.ghosh38@gmail.com,
arjia_2005@yahoo.com, akonar@etce.jdvu.ac.in

R. Janarthanan
Dept. of Information Technology
Jaya Engineering College
Chennai, India
srmjana_73@yahoo.com

*Abstract*— The present work provides a new approach to solve the well-known multi-robot co-operative box pushing problem as a multi objective optimization problem using modified Multi-objective Particle Swarm Optimization. The method proposed here allows both turning and translation of the box, during shift to a desired goal position. We have employed local planning scheme to determine the magnitude of the forces applied by the two mobile robots perpendicularly at specific locations on the box to align and translate it in each distinct step of motion of the box, for minimization of both time and energy. Finally the results are compared with the results obtained by solving the same problem using Non-dominated Sorting Genetic Algorithm-II (NSGA-II). The proposed scheme is found to give better results compared to NSGA-II.

## I. INTRODUCTION

Co-operation is an important issue in designing multi-agent systems. It is primarily targeted to design and execute a complex task or plan by more than one simple robot instead of a powerful and much more sophisticated robot. Some works regarding co-operation is reported in [1], [2], [3], [4], [5]. The box pushing problem is stated as: given an arbitrary rigid polyhedral environment, we have to determine a continuous collision-free path for transportation of the box from a given starting point to a fixed final (goal) point [5].

A specific version of the Box-pushing problem, where two similar robots have to plan the trajectory of motion of the box from a pre-defined starting position to a fixed goal position in a given environment, containing a static number of obstacles is considered in this work [6]. The robots are capable of shifting a large box from initial position to the final goal position. The box shifting includes two basic operations: turning and translation. Turning involves both push and pull operations where translation involves only push operation. In either case, both the robots stand in one side of the box and apply forces perpendicularly to it. Sufficient spacing between the box and obstacle needs to be maintained during turning and translation of the box which is assured by adding a penalty function to the energy objective function.

The problem of box-pushing is a Multi-objective optimization problem as the primary objectives refer to the minimization of energy and time. To ensure minimum time constraint, the forces applied on the box should be maximized and on the contrary, minimum energy consumption requires minimum forces to be applied by the robots. Clearly the requirements are conflicting and there is a trade-off between these two objectives. The optimization problem has been solved here using the well-known and popular multi-objective optimization algorithm namely modified MOPSO proposed by Coello Coello et al [7].

Particle Swarm Optimization is a population based heuristic optimization algorithm, inspired by the social behavior of flocking of birds and schooling of fishes. It has been successfully used for optimizing high dimensional complex functions in continuous domain (mainly). Pareto dominance is incorporated into particle swarm optimization (PSO) in order to allow this heuristic to handle problems with several objective functions. A special mutation operator is also incorporated to enrich the exploratory capability of the algorithm.

## II. MULTI-OBJECTIVE PARTICLE SWARM OPTIMIZATION

Kennedy and Eberhart [8] proposed an approach called PSO, which was inspired by the choreography of a bird flock. The approach can be seen as a distributed behavioral algorithm that performs (in its more general version) multidimensional search [9]. There have been several recent proposals to extend PSO to handle multi-objective problems. The primary motivation of multi-objective evolutionary algorithm is to obtain Pareto-optimal solution in a single run. One of the most popular PSO approach to handle MO problem is proposed by A. Coello Coello.[7]

*Definition 1; General Multiobjective Optimization Problem (MOP):* Find the vector $\vec{x}^* = [x_1^*, x_2^*, \ldots, x_n^*]^T$ which will satisfy the m inequality constraints
$$g_i(\vec{x}) \geq 0, i = 1, 2, \ldots, m$$
the p equality constrains
$$h_i(\vec{x}) = 0, i = 1, 2, \ldots, p$$
and will optimize the vector function
$$\vec{f}(\vec{x}) = [f_1(\vec{x}), f_2(\vec{x}), \ldots, f_k(\vec{x})]^T$$
where $\vec{x} = [x_1, x_2, \ldots, x_n]^T$ is the vector of decision variables.

*Definition 2; Pareto Optimality:* A point $\vec{x}^* \in \Omega$ is Pareto optimal if for every $x \in \Omega$ and $I = \{1, 2, \ldots, k\}$ either
$$\forall_{i \in I} (f_i(\vec{x}) = f_i(\vec{x}^*))$$
or, there is at least one such that
$$f_i(\vec{x}) > f_i(\vec{x}^*)$$

*Definition 3; Pareto Dominance:* A vector $\vec{u} = (u_1, u_2, \ldots, u_k)$ is said to dominate $\vec{v} = (v_1, v_2, \ldots, v_k)$ (denoted by $\vec{u} \leq \vec{v}$) if and only if u is partially less than v, i.e.,
$$\forall i \in \{1, 2, \ldots, k\}, u_i \leq v_i \cap \exists i \in \{1, 2, \ldots, k\} : u_i < v_i$$

Definition 4; Pareto Front: For a given MOP $\vec{f}(x)$ and Pareto optimal set $PF^*(x)$, the Pareto front is defined as
$$PF^* := \{\vec{u} = \vec{f} = (f_1(x), \ldots, f_k(x)) \mid x \in P^*\}$$
where $P^*$ is the Pareto Optimal Set.

The analogy of PSO with evolutionary algorithms makes evident the notion that using a Pareto ranking scheme [10] could be the straightforward way to extend the approach to handle multi-objective optimization problems. The historical record of best solutions found by a particle (i.e., an individual) could be used to store non-dominated solutions generated in the past. The use of global attraction mechanisms combined with a historical archive of previously found non-dominated vectors would motivate convergence toward globally non-dominated solutions.

*Pseudo Code:*
```
% MAX=population size,
% POS=position of particle,
% VEL=velocity of particle,
% FIT=evaluate population (POS),
% PBEST=personal best of a particle,
% REP=non-dominated vector (particle solution)
w = 0.4;
R1 = rand(0,1);
R2= rand(0,1);

FOR i=1:MAX     initialize POS(i);
FOR i=1:MAX     VEL(i) = 0;
FOR i=1:MAX     FIT(i) = evaluate(POS(i));
FOR i=1:MAX     PBEST(i) = POS(i);
REP = non_dominated(POS(i));
Initialize GRID in objective functions space;
Locate REP particles in the hypercubes (generated by GRID);

WHILE iter<max_cycle

        Assign fitness of a hypercube that is inversely
proportional to the REP particles it containing;
        FOR i=1:MAX

                Roulette-wheel selection of a hypercube;
                GBEST = Randomly select a REP particle
on that hypercube;
                VEL(i) = w*VEL(i)+R1*(PBEST(i)-
POS(i))+R2*(GBEST-POS(i));
                POS(i) = POS(i)+VEL(i);
                IF particle go beyond search space randomly
                initialize it within search space or assign
                VEL(i) = -VEL(i);

        END FOR

        Update REP;
        Update hypercubes;
        IF REP reaches max_limit particles located in less
        populated areas of objective space are given priority
        over those lying in highly populated regions;

        FOR i=1:MAX

                IF evaluate(POS(i)) dominated over
        evaluate(PBEST(i))
                        PBEST(i) = POS(i);
                ELSEIF evaluate(POS(i)) is dominated by
        evaluate(PBEST(i))
                        Retain previous PBEST(i)
                ELSE
                        PBEST(i) =
                select_randomly(POS(i),PBEST(i));
                END IF

        END FOR

        iter =iter+1;

END WHILE
```

*Use of a Mutation Operator:*
PSO is known to have a very high convergence speed. However, such convergence speed may be harmful in the context of multi-objective optimization, because a PSO-based algorithm may converge to a false Pareto front (i.e., the equivalent of a local optimum in global optimization). The RPSO [10] resets the position of a specific particle, at a certain (fixed) number of iterations. However, this approach is not only adding exploratory capabilities to PSO, but it also ensures that the full range of every decision variable is explored. The mutation rate is gradually decreased over time. Here we use a nonlinear exponentially decay function to implement the fact.

% particle = particle to be mutated
% dim = number of dimensions
% current_gen = current iteration
% tot_gen = total number of iterations
% mut_rate = mutation rate

```
IF flip((1-currentgen/totgen)^(5/mut_rate))
    which_dim = rand(0,dim-1);
    mutrange = (upperbound[which_dim]-lowerbound[which_dim])* (1-currentgen/totgen)^(5/mut_rate) ;
    ub = particle[which_dim]+mutrange;
    lb = particle[which_dim]-mutrange;
    IF lb < lowerbound[which_dim]    lb = lowerbound[which_dim];
    IF ub> upperbound[which_dim]    ub = upperbound[which_dim];
    Particle[whichdim] = RealRandom(lb,ub);

END IF
```

### III. PROBLEM FORMULATION

Consider the situation where two robots $R_1$ and $R_2$ work co-operatively to push a box from a given initial position to the goal position. Let the robots apply forces at point E $(x_e, y_e)$ and F $(x_f, y_f)$ respectively on a rectangular box ABCD, whose front wall is AD, current centre of gravity $(x_c, y_c)$ and the centre of gravity of the final goal position is $(x_{cg}, y_{cg})$ as shown in Fig. 1. The box will first turn and then move by the forces applied by the two robots. If the box turns around a point on the edge EF in anticlockwise direction, $R_1$ pulls and $R_2$ pushes the box. For clockwise rotation, the robots change their role.

Let α be the angle of rotation and I $(x_I, y_I)$ be the point around which the box is rotated. After rotation the centre of gravity becomes $(x_{cnew}, y_{cnew})$ and the new positions of the robots are $(x_{enew}, y_{enew})$ and $(x_{fnew}, y_{fnew})$.
From principle of kinematics,

$$x_{cnew} = x_I(1-\cos\alpha) + x_c \cos\alpha - \sin\alpha(y_c - y_I) \quad (1)$$

$$y_{cnew} = y_I(1-\cos\alpha) + y_c \cos\alpha - \sin\alpha(x_c - x_I) \quad (2)$$

$$x_{enew} = x_I(1-\cos\alpha) + x_e \cos\alpha - \sin\alpha(y_e - y_I) \quad (3)$$

$$y_{enew} = y_I(1-\cos\alpha) + y_e \cos\alpha - \sin\alpha(x_e - x_I) \quad (4)$$

$$x_{fnew} = x_I(1-\cos\alpha) + x_f \cos\alpha - \sin\alpha(y_f - y_I) \quad (5)$$

$$y_{fnew} = y_I(1-\cos\alpha) + y_f \cos\alpha - \sin\alpha(x_f - x_I) \quad (6)$$

After turning by an angle α (counterclockwise), the box moves with an alignment angle θ with the x-axis.

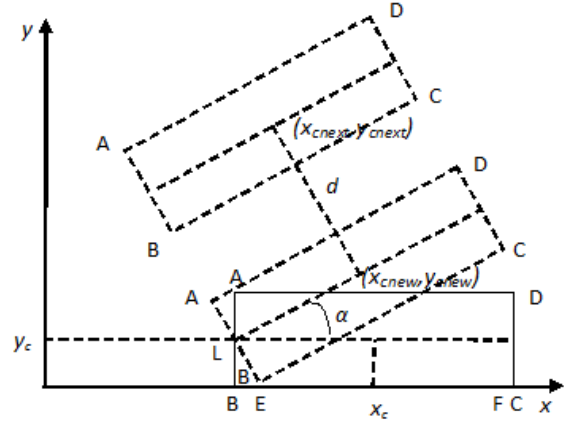

Fig. 1. Position of the box after rotation and after translation (next position)

The box moves distance $d$ with alignment angle θ and the next centre of gravity is $(x'_c, y'_c)$ where,

$$x'_c = x_{cnew} + d\cos\theta \quad (7)$$

$$y'_c = y_{cnew} + d\sin\theta \quad (8)$$

The first objective function concerns minimization of time, which has three components. The first one refers to the time required for rotation, given by

$$t_1 = \sqrt{\frac{2\alpha J}{T}} \quad (9)$$

Where, $J$ is mass moment of inertia

$T = Torque = F_{1r}d_1 + F_{2r}d_2 = 2F_{1r}d_1$ (pure rotation)
And $F_{1r}$ is the force applied by $R_1$ \quad (10)
$F_{2r}$ is the force applied by $R_2$
$d_1$ and $d_2$ are the perpendicular distance from the rotational axis to the line of action of the forces.

The second time component refers to the time needed for translation of the box to the next position. The third component refers to the predicted time cost requirement for transportation of the box from next position to the final position. Accordingly,

$$t_2 = \sqrt{\frac{2md}{F_{1t} + F_{2t}}} \quad (11)$$

Where, $F_{1t}$ is the force applied by $R_1$ to transport the box and $F_{2t}$ force applied by $R_2$ to transport the box
For translation, $F_{1t} = F_{2t}$

The third component refers to the secondary objective,

which may be added to the primary objective functions (first two components) to strengthen the quality of solutions from the point of view of other metrics. When a primary objective function gives equal measure to multiple trial solutions, the secondary objective function yields different measure for the individuals from the trial solutions. This is the significance of the secondary objective function. In present case the time needed transportation of the box from next position to the final position is treated as secondary objective.

Let $S$ be the distance between the next center of gravity and the final goal position of the center of gravity,

$$S = \sqrt{(x_c' - x_{cg})^2 + (y_c' - y_{cg})^2} \qquad (12)$$

or,

$$S = \sqrt{(x_{cnew} + d\cos\theta - x_{cg})^2 + (y_{cnew} + d\sin\theta - y_{cg})^2}.$$

The secondary objective is,

$t_3 \propto \sqrt{S}$

or, $t_3 = k\sqrt{S}$, k is a constant  (13)

Finally the first objective function is,

$f_1 = t_1 + t_2 + t_3$  (14)

The second objective function is concerned with minimization of energy. It has four components, energy for rotation, energy for translation, the secondary objective energy function and the penalty function corresponding to energy objective. Let $E_1$ and $E_2$ represent energy for rotation and translation respectively,

$E_1 = T\alpha = 2F_{1r}d_1\alpha$  (15)
$E_2 = (F_{1t} + F_{2t})d = 2F_{1t}d$  (16)

The secondary objective,

$E_3 \propto S$

$E_3 = k_1 S$, $k_1$ is a constant  (17)

In the process of selecting next position of the box from its current position, special care should be taken in order to ensure that the next position is not in close vicinity of obstacles/sidewalls of the robots workspace. This is given by the penalty function. The penalty function has a large value when the next position of the box is close enough to an obstacle. The penalty function has low value when the next position is away from the obstacle or sidewall of the world map. The penalty function is given by,

$E_4 = penalty = 2^{-d_2} k_2$  (18)

Where $k_2$ is a constant, and $d$ is a function of distance of the box with the obstacles and sidewalls, and is measured as

$d_2 = \min(d_{w1}, d_{w2}) + \min(d_{l1}, d_{l2}) + \min(d_{w3}, d_{w4})$  (19)

Where $d_{w1}$, $d_{w2}$, $d_{l1}$, $d_{l2}$, $d_{w3}$ and $d_{w4}$ are the distances of the vertices of the box with the sidewall of the workspace. The pictorial representation is given in Fig. 2.

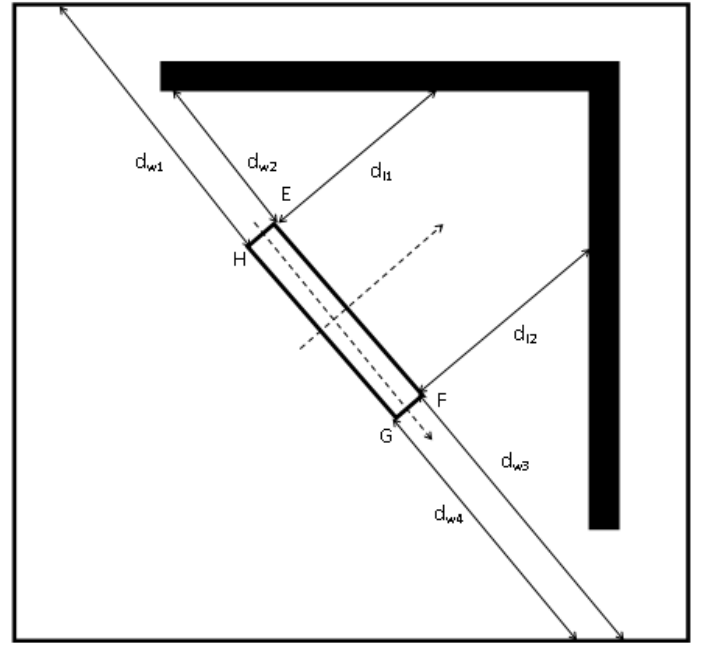

Fig. 2. Diagram for calculating d.

Finally the energy objective,

$f_2 = E_1 + E_2 + E_3 + E_4$  (20)

Here the objectives are function of $x_l$, $y_l$, $F_{1r}$, $F_{1t}$, $d_1$, $d$ and $\alpha$ which have to be determined to optimize the objective functions.

Now a solution is proposed to the problem, which presumes current center of the box, and determines the forces to be applied by the two robots to the box to shift it to the next position of the CG of the box.

*The Pseudo Code:*
Input: Initial CG of the box $(x_c, y_c)$, Final CG of the box $(x_{cg}, y_{cg})$, points of application of the two forces on the box by the two robots $(x_e, y_e)$ and $(x_f, y_f)$ and a threshold value $\varepsilon$
Output: $F_{1r}$, $F_{1t}$, $d_1$, $d$ and $\alpha$ for each step and total energy and time spent for the entire job
Begin:
Set: $x_{curr} \leftarrow x_c$; $y_{curr} \leftarrow y_c$;
Repeat
    Call MOPSO ($x_{curr}$, $y_{curr}$, $x_e$, $y_e$, $x_f$, $y_f$; $x_l$, $y_l$, $F_{1r}$, $F_{1t}$, $d_1$, $d$, $\alpha$);
    Move-to ($x_{curr}$, $y_{curr}$);
Until $\|curr - G\| \leq \in$

// $curr = (x_{curr}, y_{curr}), G = (x_{cg}, y_{cg})$ //
End
MOPSO pseudo code is described earlier in section II.

## IV. RESULTS

Fig 3,4 show the trajectory of the box when shifted from initial to the fixed final or goal position. It is apparent that the box will first turn and then translate.

Pareto-front represents the contour of optimal solutions of the objective function. We cannot use all of them in practice for application. Therefore a single solution is needed. For this, the fitness function of the solutions is normalized and the solution for which the sum of the normalized fitness value is minimum is selected.

The simulation process takes 10-12 steps depending upon the arena. On each arena the simulation is run 10 times (for each algorithm) and the average energy and time corresponding to each step is tabulated in the table and the results are compared with the result obtained by using NSGA-II algorithm [11].

From the results, it is demonstrated that modified MOPSO infers better results than NSGA-II.

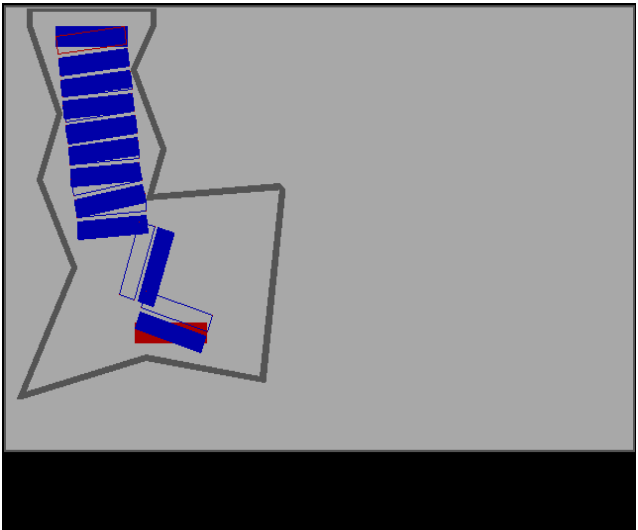

Fig. 3 Trajectory in World Map 1

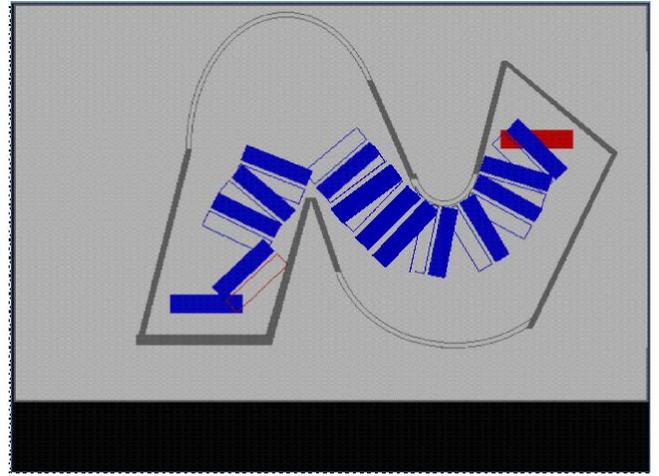

Fig. 4. Trajectory in World Map 2

TABLE I  STEP BY STEP ENERGY AND TIME FOR MOPSO ( WORLD MAP 1)

| Step | Average Energy (J) | Average Time (s) |
|---|---|---|
| 1 | 7011.45 | 56.40 |
| 2 | 6480.34 | 55.59 |
| 3 | 1518.45 | 64.58 |
| 4 | 5419.66 | 56.74 |
| 5 | 4866.35 | 62.88 |
| 6 | 1130.82 | 65.70 |
| 7 | 3672.12 | 74.13 |
| 8 | 1078.91 | 58.90 |
| 9 | 1902.95 | 73.10 |
| 10 | 221.26 | 81.49 |

TABLE II . STEP BY STEP ENERGY AND TIME FOR MOPSO  (WORLD MAP 2)

| Step | Average Energy (J) | Average Time (s) |
|---|---|---|
| 1 | 10103.99 | 30.76 |
| 2 | 2087.61 | 70.43 |
| 3 | 3880.57 | 50.99 |
| 4 | 6803.96 | 63.62 |
| 5 | 1409.06 | 76.55 |
| 6 | 5045.05 | 67.25 |
| 7 | 4957.66 | 51.15 |
| 8 | 1749.32 | 52.24 |
| 9 | 4577.89 | 64.97 |
| 10 | 950.44 | 74.62 |
| 11 | 4504.77 | 55.23 |
| 12 | 3842.23 | 83.28 |
| 13 | 2974.29 | 69.27 |

TABLE III. COMPARISON BETWEEN NSGA-II AND MOPSO

| World map | Method | Total Energy (KJ) | Total Time (s) | Total No of steps |
|---|---|---|---|---|
| 1 | NSGA-II | 38.115 | 696.57 | 11 |
| 1 | MOPSO | 33.302 | 649.51 | 10 |
| 2 | NSGA-II | 56.235 | 857.94 | 13 |
| 2 | MOPSO | 52.886 | 810.36 | 13 |

## V. CONCLUSIONS

The paper studies the scope of handling Box-pushing problem as a multi-objective optimization problem and uses the Pareto-optimal solutions to the problem using modified Multi-objective Particle Swarm Optimization. Further the results are compared with the solutions obtained by NSGA-II algorithm [6]. It has been found that MOPSO offers better and fruitful results in comparison to NSGA-II. The merit of the work lies in online optimization of the secondary objective function, which ultimately minimizes the traversal time and energy consumptions. Very often the calculation of the penalty function depends on noisy sensor data and the objective functions also become noisy. To tackle the situation, dynamic optimizing algorithms may be used. Use of Dynamic MOPSO to solve such problem would be our future endeavors.

REFERENCES

[1] Saurav Bhaumick, Indrani Goswami (Chakroborty), Amit Konar, Ananda S. Chowdhury, "Multi-robot Cooperative Box-pushing with Noisy Sensory Data Using Dynamic Multi-objective Optimization Technique" (communicated)
[2] C. R. Kube and H. Zhang (1996) "The use of perceptual cues in multi-robot box pushing", IEEE International Conference on Robotics and Automation, vol. 3, pp. 2085-2090.
[3] H. Sugie, Y. Inagaki, S. Ono, H. Aisu and T. Unemi (1995) "Placing objects with multiple mobile robots-mutual help with intension inference", IEEE International Conference on Robotics and Automation, pp. 2181-2186.
[4] Y. Yamauchi, S. Ishikawa, N. Uemura and K. Kato (1993) " On cooperative conveyance by two mobile robots", IEEE International Conference on Robotics and Systems, pp. 1478-1481.
[5] A. Verma, B. Jung and G. S. Sukatme (2001) " Robot Box-Pushing with Environment Embedded Sensors", Proceedings of 2011 IEEE international Symposium on Computational Intelligence on Robotics and Automation.
[6] Chakraborty, J.; Konar, A.; Nagar, A.; Das, S, "Rotation and translation selective Pareto optimal solution to the box-pushing problem by mobile robots using NSGA-II" Proceedings of IEEE Congress on Evolutionary Computation, 2009. CEC '09 pp- 2120 - 2126.
[7] C. A. Coello, G. T. Pulido and M. S. Lechuga, "Handling Multiobjectives with Particle Swarm Optimization", IEEE Transactions on Evolutionary Computation, 2004, vol. 3, pp. 256-279.
[8] J. Kennedy and R. C. Eberhart, Swarm Intelligence. San Mateo, CA: Morgan Kaufmann, 2001.
[9] F. van den Bergh, "An analysis of particle swarm optimization," Ph.D. dissertation, Faculty of Natural and Agricultural Sci., Univ. Petoria, Pretoria, South Africa, Nov. 2002.
[10] D. E. Goldberg, Genetic Algorithms in Search, Optimization and Machine Learning. Reading, MA: Addison-Wesley, 1989.
[11] K. Deb, A. P. S. Agarwal and T. Meyarivan,(1988) "A fast and elitist multi-objective genetic algorithm: NSGA II", IEEE Transaction on Evolutionary Computation, vol. 2, pp. 162-197